# Patchwise object tracking via structural local sparse appearance model

Hossein Kashiyani, Shahriar B. Shokouhi
School of Electrical Engineering, Iran University of Science and Technology
Tehran, Iran
hossein_kashiyani@elec.iust.ac.ir, bshokouhi@iust.ac.ir

*Abstract*—In this paper, we propose a robust visual tracking method which exploits the relationships of targets in adjacent frames using patchwise joint sparse representation. Two sets of overlapping patches with different sizes are extracted from target candidates to construct two dictionaries with consideration of joint sparse representation. By applying this representation into structural sparse appearance model, we can take two-fold advantages. First, the correlation of target patches over time is considered. Second, using this local appearance model with different patch sizes takes into account local features of target thoroughly. Furthermore, the position of candidate patches and their occlusion levels are utilized simultaneously to obtain the final likelihood of target candidates. Evaluations on recent challenging benchmark show that our tracking method outperforms the state-of-the-art trackers.

## I. INTRODUCTION

Visual object tracking is one of the significant areas in machine vision, which can be applied to motion analysis, augmented reality, vehicle navigation, and robotics. Although numerous tracking approaches have been proposed over last decade and have also achieved successes [2], [3], [5], [7], [8], [14], it is an open research problem. Object tracking depends on several challenging factors, such as motion blur, background clutter, occlusion, fast motion, out of view, scale variation, shape deformation and so on. Video object tracking approaches are approximately divided into discriminative and generative approaches. In discriminative methods [1], [3], [5], [6], a classifier differentiates object of interest from background regions. By applying this classifier at different locations, we can find target location. Grabnet et al. [1] propose feature selection based on online AdaBoost which is effectual against appearance variations. In [2], Jiang et al. integrate adaptive metric into visual tracking framework. Babenko et al. [3] propose to use multiple instance learning, which can overcome the uncertainties of training data. In the training phase, instead of assigning labels to the samples, they are assigned to the sets of samples. Updating MIL classifier by these sets, makes tracker more robust in comparison to [1]. Zhang et al. [5] demonstrate that using set likelihood for feature selection is inessential and propose online discriminative feature selection. Ma et al. [6] integrate structural local sparse descriptors into the boosting-based classifier. By exploiting structural reconstruction error of each candidate as a weight which is assigned to them, classification score can be adjusted.

In generative approaches [7], [32], [14], [17], appearance model is the fundamental core of video object tracking. Object of interest is represented by this appearance model. Then, the most similar regions to appearance model are selected as the final target. In real-world conditions, target appearance changes over time and to handle this changes, the appearance model should be updated. Considering appearance variations, Ross et al. [7] propose to progressively update low dimensional subspace representation. Despite its capability to deal with changes in illumination and scale, it cannot deal with occlusion. Zhang et al. [8] propose to employ multi-task sparse representations. Imposing joint sparsity on particle representations not only takes interdependencies between particles into consideration but also takes advantage of sparse particle representation. Ma et al. [9] models target with tensor-pooled features which are obtained from local sparse codes. This appearance model not only satisfactorily distinguishes target form background, but also alleviates dimensionality. Authors in [10] integrate inverse sparse representation with new robust distance metric which can handle difficulties. Qian et al. [11] construct two dictionaries from segmented patches called image blocks and take original and latest observations into consideration. Then, the likelihood of candidate is computed by incorporating two dictionaries. In [12], Han et al. utilize spatial information of target using dictionaries which are obtained by clustering local patches. To enhance the performance of tracker, occluded patches are eliminated using mask histogram. In [13], Wang et al. employ patches with several dimensions for construction several dictionaries. They evaluate candidates by their sparse histogram. In [14] adaptive structural local appearance (ASLA) model is used to take partial and spatial information into consideration. Motivated by Ross et al. [7], incremental subspace learning method is exploited with sparse representation for updating template in [14]. They improve ASLA model in [15] by integrating fine and coarse appearance model and employing occlusion detection to preclude entering occluded pixels in the template set. Motivated by ASLA, Zhao et al. [16] employ dual-scale ASLA to extract more features and fuse pooled features to obtain candidate similarity. Zarezade et al. [17] propose patchwise joint sparse tracker. In [17], target candidates are partitioned into non-overlapping patches and relationships of target candidates are taken into account by utilizing joint sparse representation.

In spite of remarkable success in aforementioned trackers, they are not so capable of dealing with difficulties. In this paper, we present a robust tracker that employs interdependencies of target patches over time in structural local sparse appearance model. Using joint sparse representation, the interdependencies of target patches can be considered. To construct a robust local appearance model, overlapping patches are sampled with two different sizes. Consequently, two dictionaries are needed for local sparse coding, which are constructed of similar corresponding patches in all patch templates. Marginal patches usually entail background information and tend to be occluded more than central patches. To address this issue and also prevent entering occluded patches, weighting mechanism using patch-wise reconstruction error is utilized. In joint sparse representation, corresponding similar patches construct buffers. Temporal similarity assumption should be observed in these buffers. To satisfy this assumption, in contrast to [17], our algorithm restricts updating buffers by employing occlusion levels. Furthermore, to impede entering occluded patches in dictionaries in updating section, occlusion levels are employed again.

Our contributions can be summarized as follows.
- Integrating joint sparse representation into structural local sparse appearance model to consider correlation of targets over time.
- Taking advantage of robust local appearance model using two patchwise dictionaries with different patch sizes.
- Weighting patches in structural local sparse appearance model by utilizing patchwise reconstruction error and spatial information of patches.

## II. TRACKING ALGORITHM

We introduce the appearance model and update strategy in Section II-A and II-B, respectively. The flowchart of our local sparse appearance model is illustrated in the Fig. 1.

### A. Appearance Model

With knowledge of target position in the first frame, $n$ templates $T = \{T_1, T_2, ..., T_n\}$ are cropped to construct template set after normalizing target size to $32 \times 32$ pixels. To take advantage of temporal similarity assumption, for constructing the dictionaries, each template is divided into corresponding overlapped patch sets, i.e.

$$B = \left[ \overbrace{b_1^{(1)} \ldots b_n^{(1)}}^{B^1} | \ldots | \overbrace{b_1^{(M)} \ldots b_n^{(M)}}^{B^M} \right] \in \mathbb{R}^{b \times (n \times M)} \quad (1)$$

$$D = \left[ \underbrace{d_1^{(1)} \ldots d_n^{(1)}}_{D^1} | \ldots | \underbrace{d_1^{(N)} \ldots d_n^{(N)}}_{D^N} \right] \in \mathbb{R}^{d \times (n \times N)} \quad (2)$$

where $d$ and $b$ are the dimensions of vectorized patches. $d_p^{(q)}$ and $b_p^{(q)}$ are $q$th vectorized grayscale patch of $p$th template in template set. $B^q$ and $D^q$ are $q$th group of corresponding patches. $n$ denotes the number of target templates and $M$ with $N$ denote the numbers of local patches. Target patches, denoted by $r^{(i)}$ can be constructed by columns of $B^i$ and $D^i$ [17] i.e.

$$r^{(i)} = D\alpha^{(i)} \quad i = 1,2,...,N \quad (3)$$

$$r^{(i)} = B\rho^{(i)} \quad i = 1,2,...,M \quad (4)$$

where $\alpha^{(i)}$ and $\rho^{(i)}$ are corresponding sparse codes over two different dictionaries. Considering temporal similarity assumption, each patch at the specific location has similar sparsity pattern with corresponding patches in the previous candidates in template set. To enforce joint sparsity, we utilize convex $\ell_{2,0}$ mix norms to calculate the sparse coefficients of local patches [17]. We solve the following convex problem with Simultaneous Orthogonal Matching Pursuit (SOMP) algorithm [30] as:

$$\arg\min_{A^{(i)}} \frac{1}{2} \left\| R^{(i)} - DA^{(i)} \right\|_F^2 \quad s.t. \quad \left\| A^{(i)} \right\|_{2,0} \leq \Upsilon \quad (5)$$

$$\arg\min_{P^{(i)}} \frac{1}{2} \left\| R^{(i)} - BP^{(i)} \right\|_F^2 \quad s.t. \quad \left\| P^{(i)} \right\|_{2,0} \leq \Upsilon \quad (6)$$

where $\left\| X^{(i)} \right\|_{p,q} = (\sum_{i=1}^{n+d}(\|X_i\|_p)^q)^{1/q}$, $\Upsilon$ is the maximum number of atoms for solving the convex problem. $R^{(i)} = [\Psi^{(i)}, r_t^{(i)}]$ represents $i$th patch in the current frame with four corresponding patches $\Psi^{(i)}$ in previous frames in template set, called buffer. Indeed, selection of previous patches for constructing the buffer is restricted by some conditions. In order to preclude entering dissimilar patches into buffer to observe temporal similarity assumption. Ultimately, coefficient vectors of each candidate $A = [A^{(1)}, ..., A^{(N)}]$ and $P = [P^{(1)}, ..., P^{(M)}]$, which contain the coefficients of buffer patches, are obtained. Thus, current patches coefficients are extracted as:

$$\alpha^{(i)} = A \otimes \varphi^i \quad i = 1,2,...,N \quad (7)$$

$$\rho^{(i)} = P \otimes \varphi^i \quad i = 1,2,...,M \quad (8)$$

$$\varphi^i = \begin{cases} 1, & mod(5,i) = 1 \\ 0, & others. \end{cases} \quad (9)$$

$\otimes$ indicates the element-wise multiplication. Sparse coefficients of $i$th patch $\alpha^{(i)}$ and $\rho^{(i)}$ are divided into $n$ fragments, i.e., $\alpha^{(i)T} = [a_1^{(i)T}, a_2^{(i)T}, ..., a_n^{(i)T}]$, $\rho^{(i)T} = [\rho_1^{(i)T}, \rho_2^{(i)T}, ..., \rho_n^{(i)T}]$, where each fragment corresponds to a template in template set. By accumulating these fragments [14], the frequency that each region appears in the template set is considered as follows:

$$z^{(i)} = \frac{1}{c}\sum_{j=1}^{n} a_j^{(i)}, i = 1,2,...,N \quad (10)$$

$$v^{(i)} = \frac{1}{c}\sum_{j=1}^{n} \rho_j^{(i)}, i = 1,2,...,M \quad (11)$$

where $c$ is normalization constant. Local vectors of $v^{(i)}$ and $z^{(i)}$ construct square matrices, called $V$ and $Z$. Benefiting from these

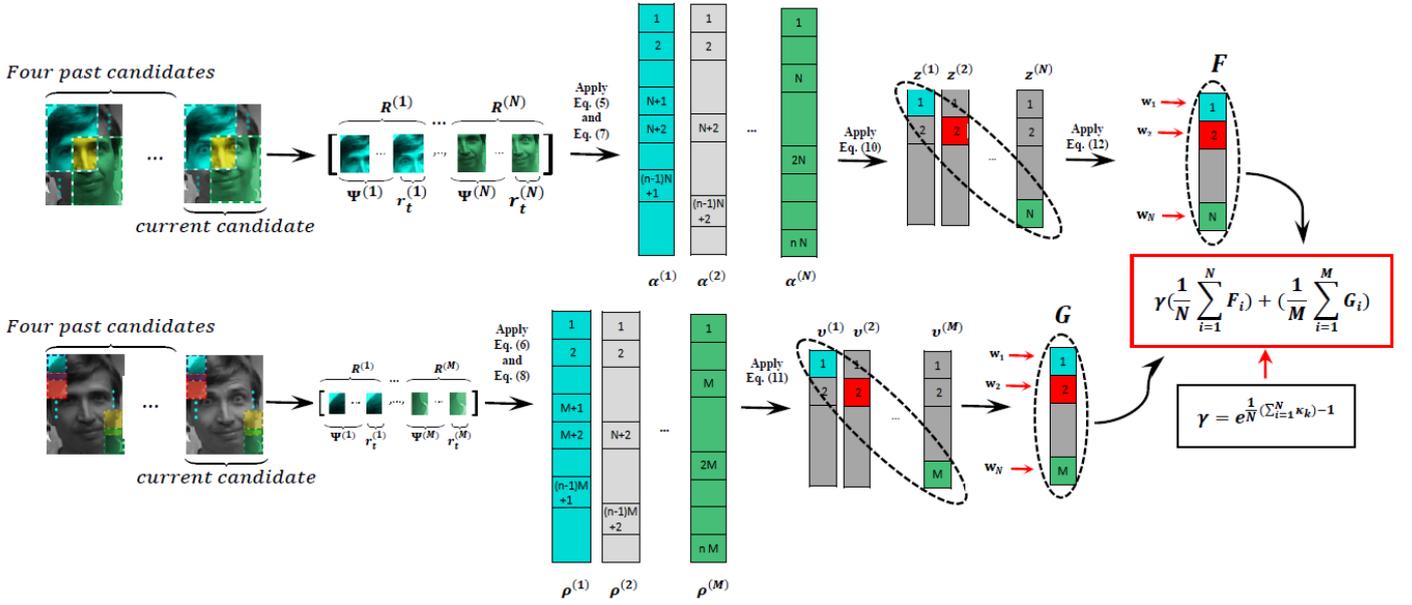

Fig. 1. Flowchart of our local sparse appearance model.

local vectors, patches that do not suffer from severe impulse noises, can be demonstrated solely by sparse codes with aligned locations. Thus, alignment pooling is employed for pooling feature of candidate, which extracts diagonal elements of square matrices, i.e., $F = diag(V)$, $G = diag(Z)$

It is noteworthy that pooled features do not take into consideration the patches position importance in candidates and their occlusion levels. Marginal regions of target candidates, which usually contain background information, are not as important as central regions. Furthermore, occluding objects initially emerge in marginal regions. Therefore, patches position is considered to enhance discriminability of appearance model. Motivated by [18], a weighting technique is proposed. In order to reduce its computational complexity for determining the occlusion levels using MIL&SVM [19], patchwise reconstruction errors are merely exploited. The weighting rule for the large-scale group of patches is formulated as:

$$\omega_k = 1 + \kappa_k e^{-\beta \left( \left| i - \frac{1+w}{2} \right| + \left| j - \frac{1+u}{2} \right| \right)}, k = 1, 2, \ldots, N$$

$$i = 1, 2, \ldots, w, j = 1, 2, \ldots, u \quad (12)$$

$$\kappa_k = \begin{cases} 1, & if \; \left\| r^{(i)} - D(\alpha^{(i)} \otimes \theta^{(i)}) \right\|_2^2 < \varepsilon \\ 0, & others. \end{cases} \quad (13)$$

$$\theta^{(i)} = \begin{cases} 1, & mod(N, i) = 1 \\ 0, & others. \end{cases} \quad (14)$$

where $w$ and $u$ are patches numbers in the row and column orientation. The weighting rule for the small-scale patch group is similar to the equation (12). Ultimately, the final weighted pooled features should be integrated together to form likelihood of target candidate. The final pooled features, which are obtained from patches with two different sizes, will have different weights in the final likelihood of target candidate.

Benefiting from the candidate reconstruction errors, a weighting method is exploited and eventually, the final likelihood of target candidate is calculated.

### B. Update Strategy

Object tracking with fixed template set tends to fail when object appearance varies to some extent due to occlusion, deformation, scale variation, and so on. To address this issue, template set should be adaptive and updated. We propose a template update method that utilizes occlusion mask for eliminating occluded patches. Instead of these occluded patches, incremental subspace representation [7] is substituted as:

$$T_n = C \otimes K + H \otimes (1 - K) \quad (15)$$

$$K = \{\kappa_1, \ldots, \kappa_N\} \quad (16)$$

where C is the best current candidate. H represents the reconstructed patches by incremental subspace learning and sparse representation, which eigenbasis vectors are updated by tracking results. As [15], a guided filter is exploited for alleviating the artifacts of fusing overlapped patches. After updating template set, dictionary can also be updated. In severe and long-term occlusion cases, dictionary updating is restricted via a condition that is defined by employing occlusion levels as:

$$\left( \frac{\sum_{k=1}^{M} \tau_k}{M} \right) + \left( \frac{\sum_{k=1}^{N} \kappa_k}{N} \right) > 2\delta \quad (17)$$

$$\tau_k = \begin{cases} 1, & if \; \left\| r^{(i)} - B(\rho^{(i)} \otimes \xi^{(i)}) \right\|_2^2 < \varepsilon \\ 0, & others. \end{cases} \quad (18)$$

$$\xi^{(i)} = \begin{cases} 1, & mod(M, i) = 1 \\ 0, & others. \end{cases} \quad (19)$$

where $\delta$ is a constant, $\tau_k$ and $\kappa_k$ are occlusion levels of the $k$th patch with $b$ and $d$ dimensions, respectively. In order to track the object of interest satisfactorily, all group components should resemble each other. In other words, tracking results rely on temporal similarity assumption in joint sparse representation. We update both groups of buffers under peculiar conditions. The corresponding conditions to large-scale and small-scale groups of buffers are written, respectively as:

$$\left(\frac{\sum_{k=1}^{N} \kappa_k}{N}\right) > o_1 \quad (20)$$

$$\left(\frac{\sum_{k=1}^{M} \tau_k}{M}\right) > o_2 \quad (21)$$

where $o_1$ and $o_2$ are constant, $\kappa_k$ and $\tau_k$ are calculated from the equations (13) and (18), respectively. Updating buffers can be controlled by determining the appropriate values for $o_1$ and $o_2$.

### III. PARTICLE FILTER

Particle filter can be seen as recursive Bayesian filtering using sequential importance sampling, which approximates the posterior probability density function of a system state. State variable of target is denoted as $u_t = \{l_x, l_y, \vartheta, s, \psi, \varphi\}$, where $l_x, l_y, \vartheta, s, \psi, \varphi$ indicate two-dimensional translations, rotation angle, aspect ratio and skew, respectively [7]. Given observations set $r_{1:t} = \{r_1, \ldots, r_t\}$ up to $t$th frame, the state variable of target is calculated by maximum a posteriori estimation as:

$$\hat{u}_t = arg\max_{u_t^i} p(u_t^i | r_{1:t}) \quad (22)$$

where $u_t^i$ denotes $i$th sample state. The posterior probability $p(u_t | r_{1:t})$ is computed as:

$$p(u_t | r_{1:t}) \propto p(r_t | u_t) \int p(u_t | u_{t-1}) p(u_{t-1} | r_{1:t-1}) du_{t-1} \quad (23)$$

where $p(u_t | u_{t-1})$ is the dynamic model and $p(r_t | u_t)$ is the observation model that indicates the likelihood of $r_t$ at the $u_t$ state. Target motion is modeled by affine transformation [7] over time and dynamic model can be estimated as $p(u_t | u_{t-1}) = N(u_t; u_{t-1}, \Sigma)$, where $\Sigma$ is diagonal covariance matrix. Eventually, the observation method is formulated as:

$$p(r_t | u_t) = \gamma(\frac{1}{N} \sum_{i=1}^{N} F_i) + (\frac{1}{M} \sum_{i=1}^{M} G_i) \quad (24)$$

where $\gamma$ is calculated as illustrated in Fig. 1. The impact of large-scale patches group in the final likelihood of target candidate is down weighted with their own reconstruction errors. In fact, using their own reconstruction errors for down weighting enhances the tracking performance.

### IV. EXPERIMENTAL RESULTS AND DISCUSSION

All experiments in this paper are simulated by MATLAB on a personal computer with Intel i7-6700HQ (2.60 GHz) and 16 GB memory. The tracking speed is roughly 1 frame per second. We utilize the SPAMS package [31] for implementation of SOMP algorithm and maximum number of atoms $\Upsilon$ is set to 4 in all image sequences. The target region is initially tagged and its bounding box is resized to $32 \times 32$ pixels. In incremental subspace learning, 20 eigenvectors are used. Template set is comprised of 10 templates and it is updated every five frames. Large-scale buffers and their corresponding dictionary are consisted of patches with size of $16 \times 16$ pixels and 8 pixels as step size. Small-scale buffers and their corresponding dictionary are consisted of patches with size of $8 \times 8$ pixels and 2 pixels as step size. Small-scale and large-scale patches in buffers and dictionaries are updated every 20 and 5 frames, respectively. The parameters $\beta, \delta, \varepsilon, o_1$ and $o_2$ are set to 0.8, 0.08, 0, and 0.13 respectively. Six parameters of affine transformation are set to [6, 6, 0.02, 0.002, 0.002, 0] and 650 particles are utilized in particle filter. The setting parameters remain unchanged during all experiments.

#### A. Evaluation On Benchmark

In order to evaluate the robustness of the presented method in dealing with wide range of difficulties, a popular benchmark [20] is used. This benchmark is composed of 51 sequences with various difficulties such as scale variation, background clutters, out of view and so on. We make the comparison of the proposed approaches with top nine trackers in the benchmark such as MIL [3], OAB [1], VTD [32], Struck [21], SCM [22], TLD [23], ASLA [14], IVT [7], CSK [24], DFT [33], VTS [25], LSK [26], CXT [27], LOT [28], and CPF [29].

#### B. Quantitative Evaluation

**Evaluation Criteria**: The robustness of the presented tracking algorithm is evaluated by distance precision (DP) and overlap success (OS) rate [20]. DP indicates the ratio of frames in which the center distance error does not exceed a predefined threshold. To rank the tracker's robustness, the error threshold is set to 20 pixels. OS rate is the ratio of frames that the overlap score exceeds a predefined threshold. The overlap score is calculated as:

$$S = \frac{|z_t \cap z_g|}{|z_t \cup z_g|} \quad (25)$$

where $z_t$ and $z_g$ are estimated and ground truth regions, respectively. $\cap$ and $\cup$ indicate the intersection and union of two areas, respectively, and $|.|$ measures the quantity of pixels in the area [20]. Also, the area under curve (AUC) score is exploited for this criteria. We utilize the conventional one-pass evaluation (OPE) to assess trackers. In all trackers, target region is initially tagged.

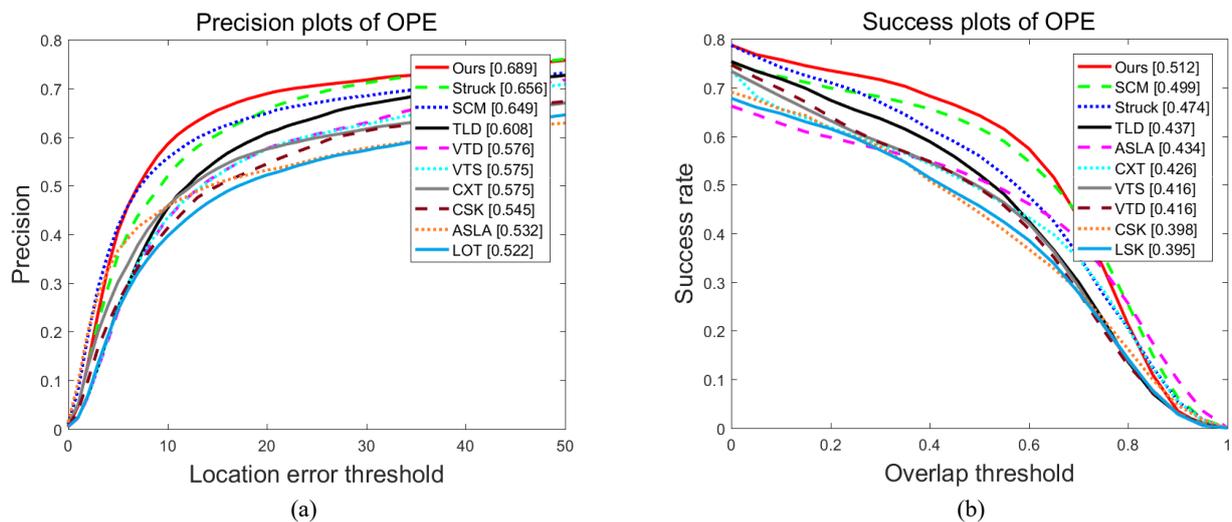

Fig. 2. Comprehensive results of precision plots (a) and success plots (b) in terms of OPE criterion. Merely top ten trackers are shown.

*1) Overall Results:* Fig. 2 shows the comprehensive results of top nine trackers on 51 sequences of benchmark in terms of DP and OS results. DP results are represented in the legend of Fig. 2(a). Moreover, the AUC scores are represented in the legend of Fig. 2(b). As shown in Fig. 2, struck tracker, SCM tracker and the proposed method obtain the best results. In the precision plots of OPE criterion, the proposed tracker achieves gains of 3.3% and 4% over Struck and SCM, respectively. In the success plots of OPE criterion, the proposed tracker achieves 1.3% and 3.8% improvement over SCM and Struck tracker, respectively. In summary, the proposed tracker achieves satisfactory results in terms of overlap precision and location accuracy.

*2) Performance Analysis per Attribute*: To evaluate the robustness of our tracking method in tackling difficulties, video sequences are labeled with 11 attributes including deformation, background clutter, scale variation, occlusion, illumination variation, out-of-plane rotation, low resolution, and so on. It is noteworthy that each video sequence can be annotated with multiple attributes. The results of the distance precision plot for 8 main challenging attributes are illustrated in Fig. 3. Fig. 3 demonstrates the capability of the presented tracking method in handling aforementioned attributes. The best improvements of distance precision in comparison with the second-best tracker relate to background clutter (7%), out-of-plane-rotation (6.8%), scale variation (4.4%), illumination variation (4.3%).

## V. CONCLUSION

To take advantage of the correlation of target patches over time, patchwise joint sparse representation for dictionary and target candidates is employed in this paper. Two dictionaries with different patch sizes are constructed to benefit from a large range of local features. To alleviate the effect of occluding object and background clutter that occur often in marginal patches, patches position and their occlusion levels are used in appearance model. In template updating, occlusion mask is exploited for eliminating occluded patches. To sum up, experimental results demonstrate the robustness of the presented tracker in dealing with deformation, background clutter, out-of-plane rotation, scale variation, occlusion, and illumination variation.

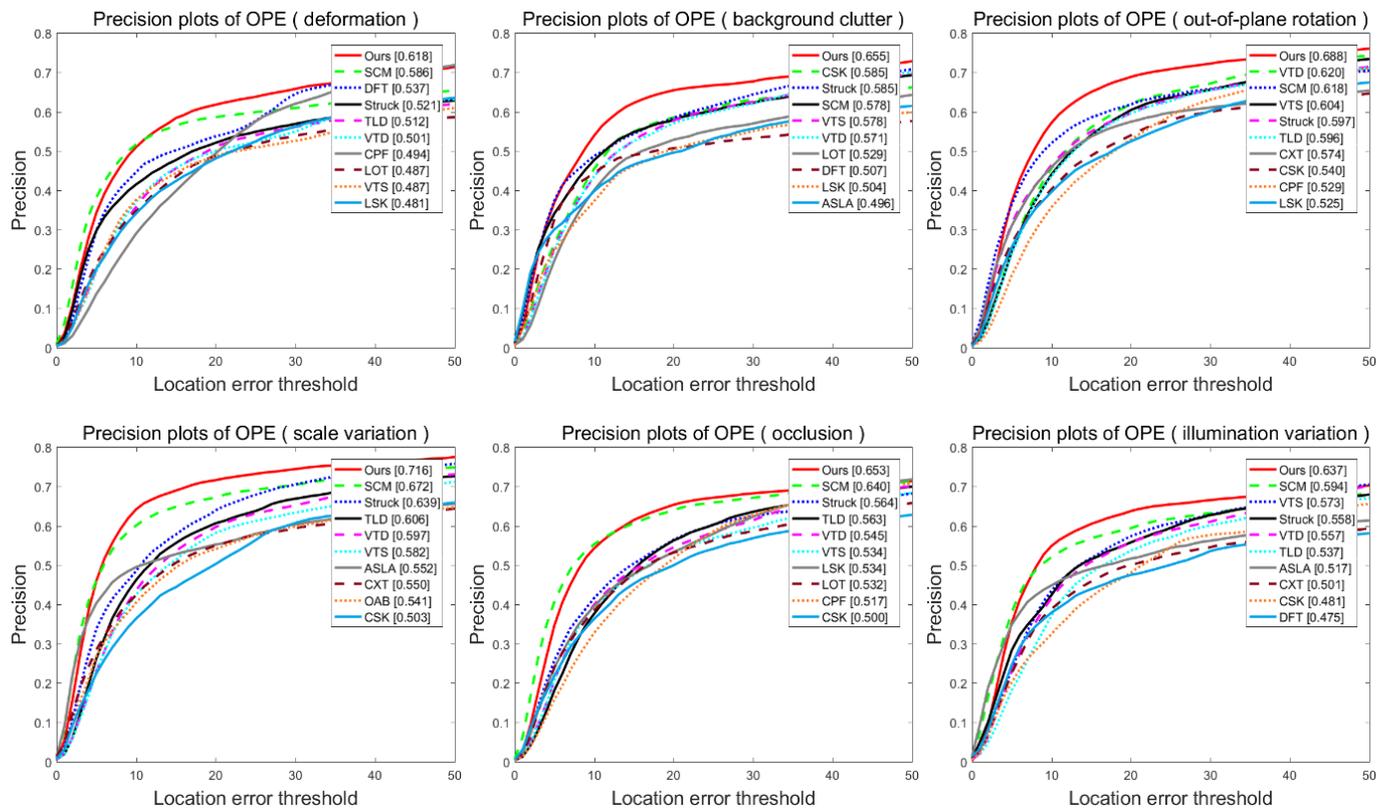

Fig. 3. Distance precision plots over six tracking challenges including deformation, background clutter, out-of-plane rotation, scale variation, occlusion, illumination variation. DP result at threshold of 20 pixels for each tracker is represented in the legend.